\documentclass[11pt]{article}

\usepackage[preprint]{acl}
\usepackage{booktabs}
\usepackage{times}
\usepackage{latexsym}
\usepackage{hyperref}

\usepackage{verbatim}
\usepackage[T1]{fontenc}
\usepackage[utf8]{inputenc}
\usepackage{microtype}

\usepackage{graphicx}
\usepackage{amsmath, amssymb, amsthm}
\usepackage{cleveref}
\usepackage{algorithm}
\usepackage[noend]{algpseudocode}
\usepackage{listings}
\usepackage{url}

\usepackage{caption}
\usepackage{subcaption}
\usepackage{linguex}
\usepackage[most]{tcolorbox}
\usepackage{multirow} 
\usepackage{tablefootnote}
\usepackage{inconsolata}

\title{In-Context Learning Boosts Speech Recognition via Human-like Adaptation to Speakers and Language Varieties}

\author{
  Nathan Roll$^{1}$, 
  Calbert Graham$^{2}$,
  Yuka Tatsumi$^{1}$,    
  Kim Tien Nguyen$^{1}$\\ 
  \textbf{Meghan Sumner}$^{1}$,
  \textbf{Dan Jurafsky}$^{1}$ \\[1ex]
  $^{1}$Stanford University\hspace{0.3cm}
  $^{2}$University of Cambridge\\[0.5ex]
  \texttt{nroll@stanford.edu}
}

\begin{document}
\maketitle

\begin{abstract}
Human listeners readily adjust to unfamiliar speakers and language varieties through exposure, but do these adaptation benefits extend to state-of-the-art spoken language models (SLMs)? We introduce a scalable framework that allows for in-context learning (ICL) in Phi-4 Multimodal (Phi-4-MM) using interleaved task prompts and audio-text pairs, and find that as few as 12 example utterances ($\sim$50 seconds) at inference time reduce word error rates by a relative 19.7\% (1.2 pp.) on average across diverse English corpora. These improvements are most pronounced in low-resource varieties, when the context and target speaker match, and when more examples are provided---though scaling our procedure yields diminishing marginal returns to context length. Overall, we find that our novel ICL adaptation scheme (1) reveals a similar performance profile to human listeners, and (2) demonstrates consistent improvements to automatic speech recognition (ASR) robustness across diverse speakers and language backgrounds. While adaptation succeeds broadly, significant gaps remain for certain varieties, revealing where current models still fall short of human flexibility. We release our prompts and code on GitHub\footnote{\fontsize{8}{12}{\url{https://github.com/Nathan-Roll1/ASR-Adaptation/}}}

\end{abstract}

\section{Introduction}
\label{sec:intro}

\begin{figure*}[htbp]
    \centering
  
    \includegraphics[width=451px, viewport=0 90 862 400, clip]{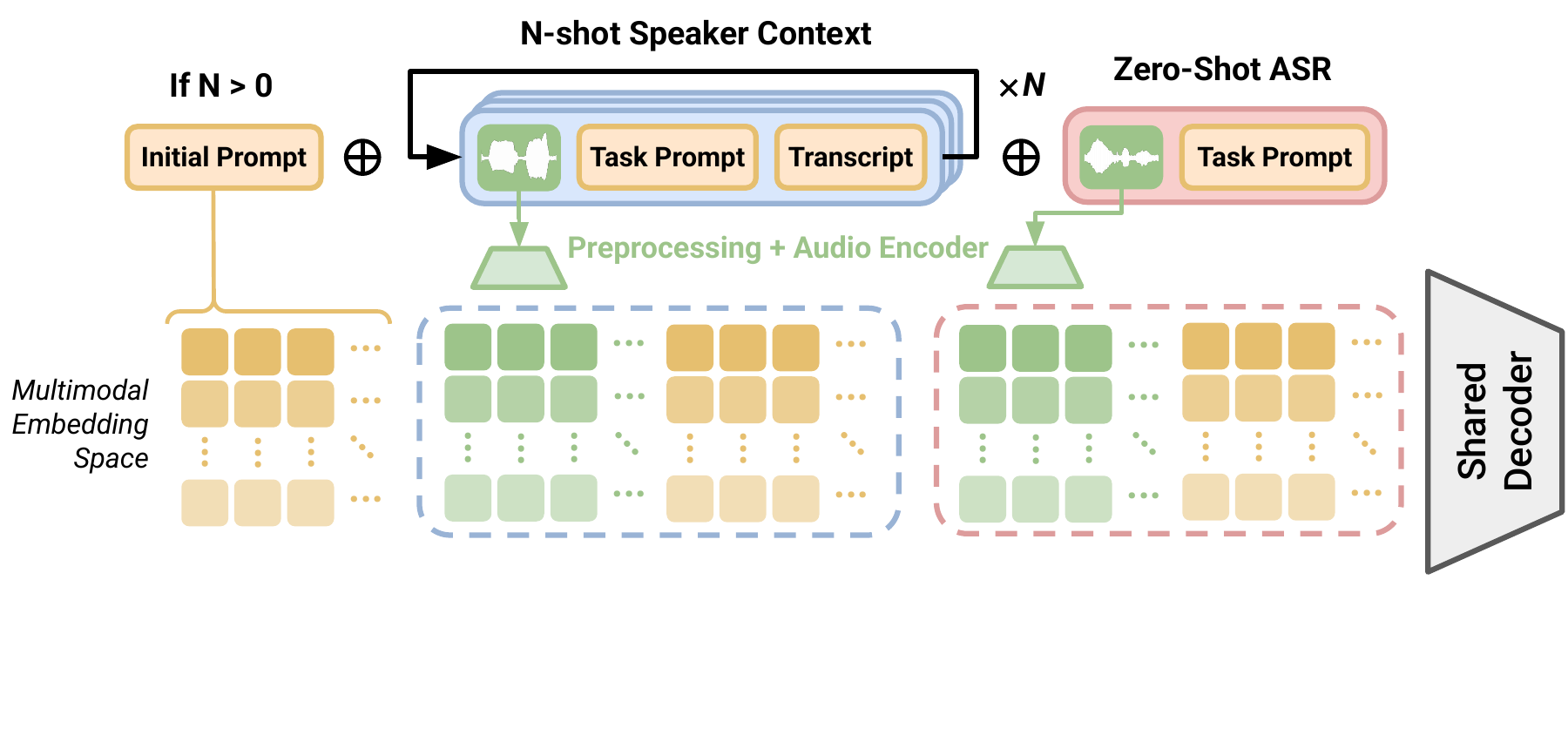} 
     \caption{Our framework provides an initial description along with \textbf{\textit{N}} transcribed examples (blue) before tasking the model to transcribe the final ASR objective audio (red). Phi-4-MM interleaves text (orange) with audio (green). These are projected into a multimodal embedding space, the context window of the shared decoder.}
    \label{fig:asr_figure}
\end{figure*}

Variation is inseparable from language---across and within accents, speakers, environments, and social settings; yet humans rapidly adapt at every level. This adaptability persists even when linguistic content is unpredictable; the mechanism is thought to involve fast (few-trial) re-weighting of acoustic--phonetic cues and recalibration of lexical priors \cite{sumner2011role, idemaru2014specificity}.

Automatic speech recognition (ASR) systems, in contrast, struggle whenever the test speaker, variety, or recording conditions diverge from the supervised training distribution. For example, word error rates (WERs) increase significantly in ``non-standard" English varieties relative to high-resource, unmarked settings \citep{rogers_effects_2004, ji_perception_2014,graham2024evaluating}. Traditional remedies, such as continued pre-training or supervised fine-tuning, are computationally expensive, cognitively implausible, and require often infeasible quantities of data \citep{azeemi2022representative, nowakowski2023adapting,bartelds2023making}.

State-of-the-art ASR systems have taken many forms in recent years. 
Contrastive learning-based encoder models like Wav2Vec 2.0 \cite{baevski2020wav2vec} or self-supervised models like HuBERT \cite{hsu2021hubert} have been surpassed in performance by encoder-decoder models like Whisper \cite{radford2023robust}. Most recently, a new class of spoken language models (SLMs) such as SALMONN \citep{tang2024salmonn}, Qwen-Audio-Chat \citep{chu2023qwen}, and Phi-4-Multimodal (Phi-4-MM) \citep{abouelenin2025phi} has pushed encoder-decoder performance even higher---beyond human levels in many settings \citep{patman2024speech, arora2025landscape}. Phi-4-MM, among the newest of these systems, has the capacity to enforce novel protocols, transcribe non-lexical features, and---for our purposes---interleave text and audio together in a way that facilitates text-guided audio prompting.  

In this paper, we ask two questions: (1) \emph{Can in-context learning (ICL) unlock human-like adaptation benefits in a state-of-the-art SLM?}, and (2) \emph{If so, does this lead to state-of-the-art performance across diverse speakers and language varieties?} We craft a simple ICL prompting framework (\cref{fig:asr_figure}) in which the model is first exposed to a handful of labeled audio-transcript pairs from the target speaker, followed by an unlabeled continuation to transcribe. Applying this setup to Phi-4-MM, we find that just a few priming utterances reduce WERs by 5.4--36.4\% (rel.) across four corpora spanning multiple English varieties \citep{kominek2004cmu,zhao2018l2,weinberger2011speech,byrne2014hispanic}. Our findings demonstrate that (i) in-context learning significantly enhances ASR robustness, especially for low-resource varieties; (ii) this adaptation shows dynamic speaker- and variety-specific effects that evolve with context length; and (iii) prompt design plays a crucial role in maximizing these benefits for underrepresented varieties. 

\section{Background}
\label{sec:related-work}

Previous work has established that listeners recalibrate phonetic categories after minimal exposure to systematic variation, whether induced by foreign accents, coarticulation, or idiolectal quirks \citep{bradlow_perceptual_2008,sidaras_perceptual_2009}. 

Phonetic variation is not merely a barrier to overcome but serves as a necessary resource for adaptation. \citet{sumner2011role} showed that listeners exposed to variable voice onset times (VOTs) from French-accented English speakers successfully shifted their phonetic boundaries, while those exposed to invariant VOTs did not adapt. This demonstrates that variation is beneficial—indeed necessary—for robust speech perception. \citet{moon2013learning} extended this work by showing that learned sub-lexical contrasts generalize across speakers of different non-native accents, with learned cues proving dominant enough to improve word recognition when paired with native contrasts. Work by \citet{demarneffe2013interaction} revealed that lexical frequency alone provides limited benefits—successful adaptation requires the interaction of phonetic variation with lexical context, not mere repetition.

Pre--deep learning pipelines relied on maximum a posteriori (MAP) adaptation and feature--space transforms such as fMLLR or i--vectors. Neural end--to--end models revived interest through layer--wise re--training, LHUC, and meta-learning \citep{klejch2018learning}. Yet these methods require either dozens of utterances per speaker or backpropagation at test time. More lightweight ideas use context biasing or rescoring with personalized language models, but benefits remain inconsistent across domains \cite{prabhavalkar2023end}.

Inspired by text LLM control, researchers have explored prefix tuning, adapters, and LoRA injections to steer multilingual ASR without updating the core model \cite{le2021lightweight, roll2025polyprompt}. Works such as \citet{le2021lightweight} show that a few frozen vectors per language can close the gap to full finetuning on talker--independent tasks, while scaling negligibly in parameters. However, most studies optimize on supervised validation sets and do not test zero--shot adaptation at inference.

We are not the first to provide labeled exemplars directly at inference time. Early sequence--to--sequence ASR treated preceding audio--transcript pairs as an additional context window \cite{kim2023sgem}. Whisper's dense logits make such prompting tricky, but \citet{wang2024can} and \citet{chen_salm_2024} independently showed sizable WER drops by concatenating audio--text pairs, especially for dialectal Chinese. Later, COSMIC introduced instruction tuning to reinforce the format, while Phi-4-MM extends the paradigm to low--footprint models. Our work focuses on the specific schema for implementing ICL in SLMs like Phi-4-MM, detailing the interleaving of task prompts, ground truth transcriptions, and audio exemplars within a shared context window while studying the effectiveness, scaling, and cognitive plausibility of ICL.

Evidence from bilingual production suggests that talker-specific traits such as speaking rate \citep{bradlow_language-independent_2017, graham2019articulation} or tonal structure \citep{graham2018second} carry over from L1 to L2 . For ASR, cross--lingual prompts or multilingual adapter stacks can leverage high-resource L1 data to bootstrap L2 decoding \cite{hsu2024meta}. Our work intersects these lines by probing whether an English--centric SLM can nevertheless exploit talker-specific cues shared across dialects and second languages. To our knowledge, this is the first study to apply in-context learning for speaker adaptation in ASR across multiple speech corpora, and the first to apply these paradigms in multimodal language models.

\section{Data}
\label{sec:data}

Our experiments leverage four English speech corpora that collectively span diverse speaker demographics, accent varieties, and speech contexts. This selection enables comprehensive evaluation of in-context adaptation across different types of linguistic variation while maintaining experimental rigor through controlled comparisons.

\subsection{L2-ARCTIC}
\label{subsec:l2arctic}

L2-ARCTIC \citep{zhao2018l2} contains high-quality recordings from 24 non-native English speakers representing six major world languages: Hindi, Korean, Mandarin, Spanish, Arabic, and Vietnamese. Each first language group includes two male and two female speakers, providing balanced gender representation. Each speaker recorded approximately one hour of read speech consisting of 1,132 phonetically balanced sentences adapted from the CMU ARCTIC prompt set \citep{kominek2004cmu}.

\subsection{CMU-Arctic}
\label{subsec:cmuarctic}

CMU-Arctic \citep{kominek2004cmu} is comprised of approximately 18 hours of phonetically balanced American English read speech across 18 speakers. Each speaker read the same set of approximately 1,200 utterances designed for comprehensive coverage of American English phonetic contexts. While featuring primarily American English speakers, the corpus also includes speakers with German, Indian, and other regional backgrounds, providing some accent diversity within the ``native'' category.

\subsection{Hispanic-English Corpus (HEC)}
\label{subsec:hec}

The Hispanic-English Corpus \citep{byrne2014hispanic} contains approximately 30 hours of bilingual speech data from 22 Spanish heritage speakers residing in the United States. Speakers were adult native Spanish speakers from Central and South America who had lived in the United States for at least one year. For this study, we use only the English read speech portions to maintain consistency with other corpora.

\subsection{Speech Accent Archive (SAA)}
\label{subsec:saa}

The Speech Accent Archive \citep{weinberger2011speech} contains approximately 23 hours of English speech from over 2,500 speakers representing more than 200 first language backgrounds worldwide. All speakers read the identical 69-word paragraph beginning with ``Please call Stella...''. This uniform elicitation enables systematic comparison across accent types while controlling for lexical and syntactic factors. Given the identical elicitation paragraph, we utilized SAA to benchmark 0-shot ASR performance disparities across a wide range of accents within the Phi-4-MM specifically, and not to evaluate the proposed ICL framework.

\subsection{Data Selection Rationale}

These four corpora were selected to provide complementary perspectives on accent adaptation while enabling rigorous experimental control. The shared elicitation materials between L2-ARCTIC and CMU-Arctic enable direct comparison of adaptation effects for native versus non-native speakers under identical linguistic conditions. Together, the corpora span native American English, major world language varieties, Spanish heritage varieties, and global accent diversity, providing comprehensive coverage of English pronunciation variation.

For this study, we filtered speakers to ensure adequate context examples for few-shot evaluation, including only speakers with at least 13 valid utterances (minimum 2.5 seconds duration) and varieties represented by at least two speakers. This ensured that both test and context utterances could be drawn from the same variety with sufficient speech material to construct few-shot prompts of varying lengths. For each test utterance, we randomly sampled a fixed number of non-overlapping utterances from either the same speaker or a different speaker of the same variety, depending on the experimental condition. This procedure was repeated for all shot count conditions (0-12). After filtering, our analysis included 15 speakers from CMU-Arctic, 14 speakers from L2-ARCTIC, 7 speakers from HEC, and the full SAA corpus for zero-shot evaluation of the Phi-4 model.

\section{Model: Phi-4-MM}
\label{sec:model}

\textbf{Phi-4-MM} builds on a frozen Phi-4-Mini-Instruct core by integrating dedicated encoders for vision and audio via lightweight LoRA, enabling unified text generation from multimodal inputs \cite{abouelenin2025phi}.  The model supports up to 128 thousand tokens of context and generates outputs in dozens of languages.

For speech/audio, Phi-4-MM accepts 80-dimensional log-Mel filter-bank frames and processes them through a convolutional front end followed by Conformer blocks \cite{gulati2020conformer}.  A two-layer projector then maps encoded audio into the text embedding space, where modality-specific LoRA adapters interface with the frozen layers.

{\bf Pre-training} aligns the audio encoder and frozen text decoder using approximately 2 million hours of anonymized speech–text pairs spanning eight languages (Chinese, English, French, German, Italian, Japanese, Portuguese, Spanish).  This stage uses only paired ASR data to teach the model cross-modal semantic alignment.

{\bf Instruction fine-tuning} After the pre-training phase, Phi-4-MM is fine-tuned on roughly 100 million curated speech and audio samples—covering ASR, speech translation, question answering, summarization, and broader audio understanding—across the same eight languages.  Maximum audio lengths vary by task (from ~30 seconds for ASR to ~30 minutes for summarization), ensuring the model learns both short-form and long-form speech processing in diverse linguistic contexts.

\section{Methods}
\label{sec:methods}

\subsection{In-Context Learning Framework}
\label{subsec:icl_framework}

We introduce a novel prompting framework that enables Phi-4-MM to perform fast, low-data adaptation through ICL. Our approach leverages the multimodal capabilities of the model by interleaving transcribed audio-text pairs as exemplars before presenting target audio for transcription. Unlike traditional ASR adaptation methods that require parameter updates or extensive speaker-specific data, our approach achieves adaptation purely through prompt engineering at inference time.

The framework operates by providing $N$ audio-transcript example pairs (``shots'') followed by a target audio segment to be transcribed. We systematically evaluated 0 through 12 in-context examples to capture both initial adaptation effects and scaling effects. Each prompt includes a series of <|user|> audio inputs paired with <|assistant|> transcriptions, followed by an unlabeled test audio segment. Full prompt templates and token formatting are provided in Appendix \ref{app:icl_prompts}.

\begin{figure*}[!htbp]
  \centering
  \includegraphics[width=\textwidth]{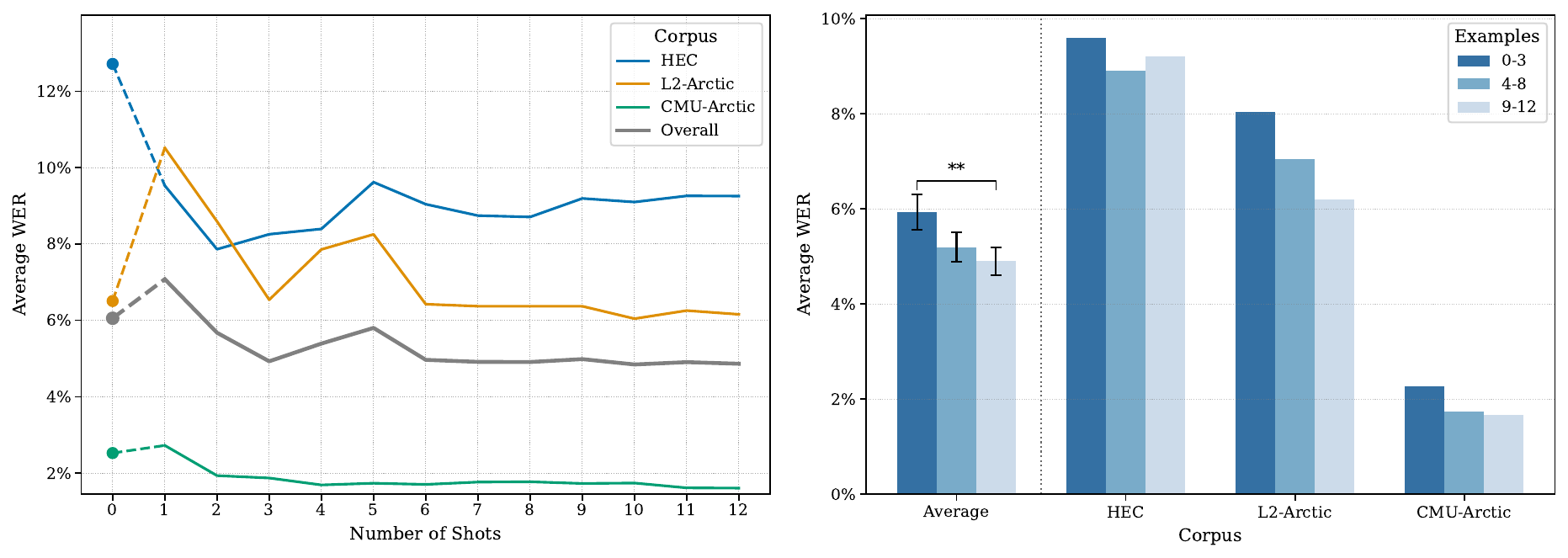}
  \caption{
  In-context learning consistently reduces WERs across all corpora with diminishing returns. (\textbf{Left}) WER trajectories by shot count show rapid initial improvement plateauing around 6-10 examples. (\textbf{Right}) Aggregated results across shot buckets (0-3, 4-8, 9-12) demonstrate strictly decreasing WERs with more examples. CMU-Arctic (native speakers) achieves lowest WERs across all conditions.
}
  \label{fig:compound_marginal_plot6}
\end{figure*}

\subsection{Prompt Design and Speaker Context Conditions}
\label{subsec:prompt_design}

We developed two prompting strategies to investigate format specificity effects. For zero-shot evaluation, we employed both a standard prompt (``Transcribe the audio clip into text'') and a variation explicitly mentioning non-native speech. Our few-shot framework follows a structured conversation format that begins with explicit instructions, includes model acknowledgment, presents each audio-transcript pair individually, and concludes with the transcription request. The variation prompt includes explicit ``Transcription:'' markers designed to provide clearer structural cues that may benefit lower-resource varieties. 

To investigate adaptation specificity, we examined two context conditions that map onto human perceptual learning paradigms: same-speaker (within-talker evidence from the identical individual as the target) and different-speaker (within-variety evidence from other speakers of the same language variety).

\subsection{Model Configuration and Preprocessing}
\label{subsec:model_config}

All experiments used Phi-4-MM with greedy decoding to ensure deterministic outputs suitable for controlled evaluation. Technical implementation details, including a minor code adjustment to the model's \texttt{num\_logits\_to\_keep} parameter to ensure correct behavior with our generation settings, are provided in Appendix~\ref{app:model_params}.

Audio preprocessing involved resampling to 16 kHz, normalization to float32 to preserve dynamic range and numerical precision during downstream processing, and filtering clips shorter than 2.5 seconds. This duration cutoff ensures coverage of the "few-trials" regime documented in human adaptation literature while maintaining sufficient acoustic information for analysis. Standard text preprocessing included lowercasing, punctuation removal, and whitespace normalization to ensure fair and consistent word error rate (WER) calculation. Comprehensive preprocessing specifications are detailed in Appendix~\ref{app:datasets_preprocessing}.

\subsection{Experimental Design and Evaluation}
\label{subsec:experimental_design}

We applied strict filtering criteria to ensure robust evaluation: varieties required at least two speakers, speakers needed at least 13 valid utterances (enabling 12-shot evaluation), and we limited analysis to 50 utterances per speaker. This was to maintain a consistent evaluation budget across speakers and prevent over-representation of any individual voice. Context examples were selected using controlled randomization with fixed seeds for reproducibility, excluding examples with identical transcripts to the test audio.

Our primary evaluation metric was WER, computed using the \texttt{jiwer} library. Results were aggregated across trial, speaker, language variety, and corpus levels to provide comprehensive analysis of adaptation effects. We conducted up to 50 trials per speaker per condition, with experiments running on NVIDIA A100 GPUs requiring approximately 8-12 hours per corpus for complete evaluation.

The experimental design systematically tested all combinations of shot counts (0-12), speaker conditions (same/different), and prompt types (standard/variation) across the four speech corpora. This comprehensive approach enables detailed analysis of how adaptation benefits vary across different linguistic populations and experimental conditions. Speaker-level results comparing 0-shot and 12-shot performance are presented in Appendix~\ref{app:experimental_design} and grouped results are shown in \cref{fig:compound_marginal_plot6}. 

\begin{table*}[htbp]
\centering

\setlength{\tabcolsep}{4pt}
\caption{Phi-4-MM performance by shot count and corpus. Our ICL method improves performance across all corpora and nearly all language varieties, with an average WER decrease of 19.7\% rel. (1.2 pp.). Zero-shot performance on SAA (\cref{subsec:saa}) highlights high-low resource discrepancies found between the other corpora.}
\fontsize{9pt}{12pt}\selectfont
\label{tab:main_results}
\begin{tabular}{lc|ccccccccccccc}
\toprule
\multirow{2}{*}{\textbf{Corpus} / \textit{Variety}} & \multicolumn{13}{c}{\textbf{\textit{N}}-shot WER} & \multirow{2}{*}{0 $\rightarrow$ 12-shot}\\
\cmidrule(lr){2-14} 
& 0 & 1 & 2 & 3 & 4 & 5 & 6 & 7 & 8 & 9 & 10 & 11 & 12 & \\
\midrule
\textbf{CMU-Arctic}        & \textbf{2.5} & \textbf{\textcolor{RedOrange}{2.7}} & \textbf{1.9} & \textbf{1.9} & \textbf{1.7} & \textbf{1.7} & \textbf{1.7} & \textbf{1.8} & \textbf{1.8} & \textbf{1.7} & \textbf{1.7} & \textbf{1.6} & \textbf{\textcolor{ForestGreen}{1.6}} & \textbf{-0.9 (-36.1\%)} \\

\hspace*{1em}\textit{German} & \textit{3.5} & \textit{\textbf{\textcolor{RedOrange}{3.6}}} & \textit{2.9} & \textit{2.6} & \textit{2.7} & \textit{2.5} & \textit{2.5} & \textit{2.8} & \textit{2.6} & \textit{2.5} & \textit{2.6} & \textit{2.3} & \textit{\textbf{\textcolor{ForestGreen}{2.3}}} & \textit{-1.2 (-35.2\%)} \\
\hspace*{1em}\textit{Indian} & \textit{3.2} & \textit{\textbf{\textcolor{RedOrange}{4.6}}} & \textit{2.1} & \textit{2.2} & \textit{2.0} & \textit{2.0} & \textit{2.0} & \textit{1.9} & \textit{2.1} & \textit{2.0} & \textit{2.1} & \textit{\textbf{\textcolor{ForestGreen}{1.8}}} & \textit{1.9} & \textit{-1.2 (-39.2\%)} \\
\hspace*{1em}\textit{U.S.}     & \textit{\textbf{\textcolor{RedOrange}{2.0}}} & \textit{1.8} & \textit{1.7} & \textit{1.8} & \textit{1.5} & \textit{1.6} & \textit{1.6} & \textit{1.7} & \textit{1.6} & \textit{1.6} & \textit{1.7} & \textit{\textbf{\textcolor{ForestGreen}{1.4}}} & \textit{1.5} & \textit{-0.4 (-21.9\%)} \\
\hspace*{1em}\textit{Other}          & \textit{\textbf{\textcolor{RedOrange}{2.1}}} & \textit{1.6} & \textit{1.6} & \textit{1.4} & \textit{1.2} & \textit{1.3} & \textit{1.2} & \textit{1.3} & \textit{1.3} & \textit{1.3} & \textit{1.2} & \textit{1.3} & \textit{\textbf{\textcolor{ForestGreen}{1.2}}} & \textit{-0.9 (-44.5\%)} \\

\midrule
\textbf{L2-Arctic}        & \textbf{6.5}  & \textbf{\textcolor{RedOrange}{10.5}} & \textbf{8.6}  & \textbf{6.5}  & \textbf{7.9}  & \textbf{8.3}  & \textbf{6.4}  & \textbf{6.4}  & \textbf{6.4}  & \textbf{6.4}  & \textbf{6.0}  & \textbf{6.3}  & \textbf{\textcolor{ForestGreen}{6.2}}  & \textbf{-0.3 (-5.4\%)}  \\

\hspace*{1em}\textit{Hindi}       & \textit{4.0}  & \textit{\textbf{\textcolor{RedOrange}{7.1}}}  & \textit{3.9}  & \textit{4.0}  & \textit{3.7}  & \textit{3.8}  & \textit{3.8}  & \textit{3.7}  & \textit{3.6}  & \textit{3.6}  & \textit{3.7}  & \textit{\textbf{\textcolor{ForestGreen}{3.4}}}  & \textit{3.5}  & \textit{-0.5 (-13.7\%)} \\
\hspace*{1em}\textit{Korean}      & \textit{4.2}  & \textit{8.5}  & \textit{14.9} & \textit{4.5}  & \textit{14.9} & \textit{\textbf{\textcolor{RedOrange}{15.2}}} & \textit{4.3}  & \textit{3.9}  & \textit{3.9}  & \textit{3.7}  & \textit{\textbf{\textcolor{ForestGreen}{3.4}}}  & \textit{3.8}  & \textit{3.9}  & \textit{-0.3 (-7.9\%)}  \\
\hspace*{1em}\textit{Mandarin}    & \textit{7.4}  & \textit{\textbf{\textcolor{RedOrange}{11.4}}} & \textit{8.3}  & \textit{7.7}  & \textit{7.6}  & \textit{8.1}  & \textit{7.7}  & \textit{7.7}  & \textit{7.7}  & \textit{7.7}  & \textit{\textbf{\textcolor{ForestGreen}{7.6}}}  & \textit{7.8}  & \textit{7.7}  & \textit{+0.3 (+3.9\%)}  \\
\hspace*{1em}\textit{Spanish}     & \textit{6.5}  & \textit{\textbf{\textcolor{RedOrange}{10.1}}} & \textit{7.0}  & \textit{6.3}  & \textit{6.1}  & \textit{6.0}  & \textit{6.2}  & \textit{6.2}  & \textit{6.3}  & \textit{6.2}  & \textit{\textbf{\textcolor{ForestGreen}{5.7}}}  & \textit{5.9}  & \textit{5.9}  & \textit{-0.6 (-9.0\%)}  \\
\hspace*{1em}\textit{Vietnamese}  & \textit{11.3} & \textit{\textbf{\textcolor{RedOrange}{17.2}}} & \textit{13.1} & \textit{11.3} & \textit{10.8} & \textit{12.6} & \textit{11.1} & \textit{11.1} & \textit{11.2} & \textit{11.4} & \textit{\textbf{\textcolor{ForestGreen}{10.6}}} & \textit{11.4} & \textit{10.7} & \textit{-0.5 (-4.9\%)}  \\

\midrule
\textbf{HEC}             & \textbf{\textcolor{RedOrange}{12.7}} & \textbf{9.5}  & \textbf{\textcolor{ForestGreen}{7.9}}  & \textbf{8.3}  & \textbf{8.4}  & \textbf{9.6}  & \textbf{9.0}  & \textbf{8.7}  & \textbf{8.7}  & \textbf{9.2}  & \textbf{9.1}  & \textbf{9.3}  & \textbf{9.3}  & \textbf{-3.5 (-27.2\%)} \\

\midrule

\textbf{SAA\footnotemark} & \textbf{4.7}  &     &     &     &    &    &    &    &    &    &     &     & \multicolumn{1}{r}{\textbf{Avg\footnotemark:}} & \multicolumn{1}{c}{\textbf{- 1.2 (-19.7\%)}}      \\
\hspace*{1em}\textit{Native} & \textit{1.2} & & & & & & & & & & & & & \multicolumn{1}{c}{} \\
\hspace*{1em}\textit{Non-Native} & \textit{11.4} & & & & & & & & & & & & & \multicolumn{1}{c}{} \\

\cmidrule[1pt]{1-2} 

\end{tabular}
\end{table*}

\section{Results}
\label{sec:results}

Our experiments demonstrate that providing in-context audio-transcript examples consistently improves ASR performance across all tested corpora, with the magnitude and pattern of improvements varying systematically across speaker populations and experimental conditions.

\subsection{In-Context Learning Effectiveness}
Figure~\ref{fig:compound_marginal_plot6} shows that in-context learning produces substantial and generally consistent improvements across all corpora with 9-12 examples significantly better than 0-3 at a 95\% confidence level (two-sample t-test). The left panel reveals characteristic diminishing returns: the largest gains occur between 0 and 1 shots, with performance improvements plateauing around 6-10 examples. CMU-Arctic consistently achieves the lowest WERs across all shot conditions, reflecting the high-resource nature of standard American English in ASR training data. Baseline (0-shot) WERs vary widely across corpora, ranging from 2.5\% (CMU-Arctic) to 12.7\% (HEC). The HEC average is heavily influenced by a single outlier speaker (Speaker 7, Table~\ref{tab:complete_speaker_results} in Appendix~\ref{app:speaker_results}) with an exceptionally high 0-shot WER of 63.9\%; excluding this speaker, the 0-shot average for the remaining HEC speakers is approximately 4.2\%. This outlier also significantly impacts the 12-shot HEC average (41.0\% WER for Speaker 7--see \cref{app:speaker_results}). Non-native speakers in SAA reached 11.4\% compared to just 1.2\% for native speakers. 

Aggregated across shot ranges, performance follows a clear hierarchy: 9-12 shots outperform 4-8 shots, which in turn outperform 0-3 shots across all corpora. The asymptotic shape indicates that approximately 25-30 seconds of transcribed audio captures most adaptation benefits available through ICL (see \cref{fig:compound_marginal_plot6}).

\subsection{Corpus-Level Performance Patterns}
\footnotetext[2]{\textit{Native} and \textit{Non-native} are distinctions which overlook the complexity of many language-learning trajectories, however they manifest the wide gaps in ASR performance.}

Table~\ref{tab:main_results} reveals several consistent patterns across corpora and varieties. Nearly all speaker groups achieve their highest WERs in the 0-1 shot conditions, with the notable exception of L2-Arctic Korean speakers who show elevated error rates before improving substantially. Most varieties reach their highest accuracies with 10-12 examples.

Low-resource varieties generally experience large absolute and relative improvements. For HEC, the substantial average absolute WER reduction reported in Table~\ref{tab:main_results} (-3.5 points from 0 to 12 shots) is largely driven by the aforementioned outlier speaker's improvement (-22.9 points). Within CMU-Arctic, speakers with German and Indian backgrounds show relative gains of 35.2\% and 39.2\% respectively (0 to 12 shots), compared to 21.9\% for US English speakers. The "Other" category (see Section~\ref{subsec:cmuarctic} for details) achieves the largest relative improvement of 44.5\%. Similarly, in L2-Arctic, most non-native varieties achieve gains, with Hindi (-13.7\%) and Spanish (-9.0\%) showing notable improvements, while Korean shows a more modest -7.9\% gain.

\footnotetext[3]{Average over speakers. (See Appendix \ref{app:speaker_results})}

A critical finding is that baseline disparities persist despite adaptation. SAA illustrates this most starkly: in zero-shot conditions, native speakers achieve 1.2\% WERs while non-native speakers reach 11.4\% WERs—a nearly 10-fold difference despite all speakers reading identical text. L2-Arctic Mandarin is the only variety that performs slightly worse (+3.9\%) at 12 shots compared to zero-shot. This small negative change falls within expected noise levels and does not contradict the overall adaptation trend. We hypothesize that bilingual interference or mismatches between orthography and pronunciation may underlie the weaker or inconsistent adaptation patterns observed in these speaker groups.

\subsection{Speaker Context Specificity}

Table~\ref{tab:same_vs_diff_shot_groups} examines whether adaptation benefits vary when using examples from the same speaker versus different speakers of the same variety. The results reveal a nuanced pattern: no difference appears in the 1-3 shot range, but same-speaker examples provide a substantial 1.1 percentage point advantage (19.6\% relative improvement) specifically when 4-6 examples are provided. This advantage disappears entirely at higher shot counts, with different-speaker examples slightly outperforming same-speaker context at 10-12 shots.

This pattern suggests two distinct adaptation mechanisms operating at different scales: initial speaker-specific acoustic calibration that benefits from idiosyncratic features, followed by variety-level adaptation that emphasizes shared phonetic patterns across speakers within accent groups.

\begin{table}[ht]
\caption{Comparison of same-speaker versus different-speaker context performance across shot groups. Same-speaker context shows a notable benefit in the 4-6 shot range.}
\label{tab:same_vs_diff_shot_groups} 
\centering
\fontsize{9.7pt}{12pt}\selectfont
\begin{tabular}{lccl}
\toprule
 & \multicolumn{2}{c}{Speaker Condition} & \\
\cmidrule(lr){2-3}
Shot Group & \textit{Same} & \textit{Different} & \textit{Same} Advantage\footnotemark \\
\midrule
1--3   & 5.6 & 5.7 & 0.1 (+1.8\%) \\
4--6   & 4.5 & 5.6 & \textbf{\textcolor{ForestGreen}{1.1 (+19.6\%)}} \\
7--9   & 4.6 & 4.5 & -0.1 (-2.2\%) \\
10--12 & 4.5 & 4.3 & \textbf{\textcolor{RedOrange}{-0.2 (-4.7\%)}} \\
\bottomrule
\end{tabular}
\end{table}
\footnotetext{The "Same Advantage" is calculated as (Different WER - Same WER). The relative percentage in parentheses is calculated as (Absolute Advantage / Different WER) * 100\%.}

\subsection{Prompt Sensitivity and Format Effects}\label{sec:prompt-effects}

\begin{figure}[!thbp]
  \centering
  \includegraphics[width=220px]{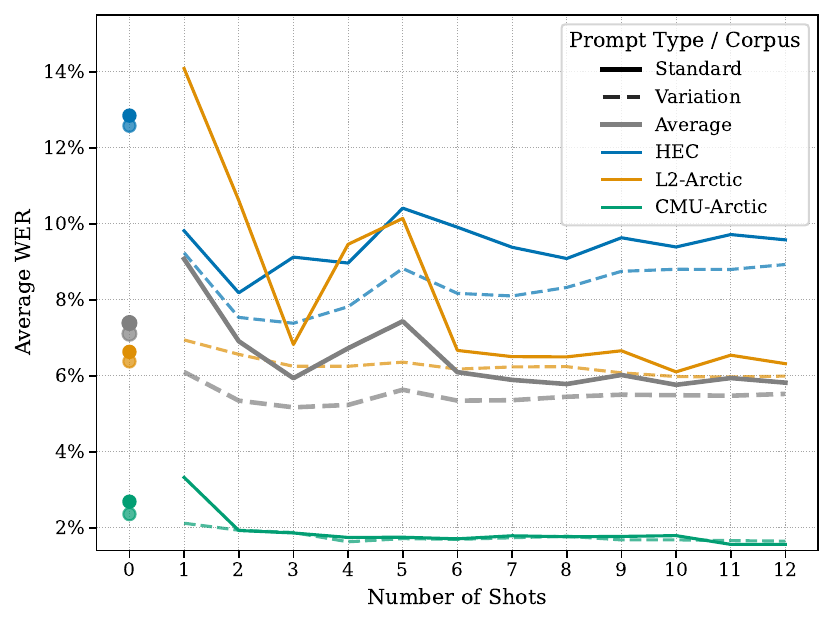}
\caption{
  Prompt format sensitivity across corpora. Including explicit ``Transcription:'' markers reduces WERs in low-resource corpora (HEC, L2-Arctic). In zero shot settings, marginal gains are induced simply by informing the model that it's transcribing non-native speech.
}
  \label{fig:prompt_stability}
\end{figure}

Figure~\ref{fig:prompt_stability} plots the impact of prompt wording on WERs.  We tested exactly \textbf{four} lightweight templates (fully specified in Appendix~\ref{app:icl_prompts}): for \emph{zero-shot} inference a \textbf{standard} instruction taken from the model card ("Transcribe the audio clip into text.") versus a \textbf{variation} that pre-labels the clip as coming from a "non-native English speaker"; for \emph{few-shot} (\textit{N}-shot) inference a standard template that simply concatenates each audio–text pair and a variation that additionally pre-generates the token "\texttt{Transcription:}" before each exemplar.  These manipulations isolate two hypothesized helpers—\emph{task framing} through explicit accent information and \emph{I/O scaffolding} through consistent answer delimiters.

In the zero-shot setting, the non-native framing yielded a \emph{small but consistent} improvement across all corpora (blue bars in Figure~\ref{fig:prompt_stability}), lowering WERs by 0.1–0.3~pp even for native speech.  The effect suggests that the phrase "non-native" activates a broader acoustic–phonetic prior learned during instruction tuning, making the decoder slightly more tolerant of unexpected phone–letter mappings.

For few-shot adaptation, injecting the "\texttt{Transcription:}" tag proved the larger lever.  It reduced early-shot volatility (<4 examples) and delivered up to 0.9pp average WER gains in L2-Arctic and 0.6pp in HEC, while leaving high-resource CMU-Arctic essentially unchanged.  Together, the two \textbf{variation} templates confirmed our a priori expectation: explicit accent cues help immediately, and explicit answer markers help the model exploit sparse context more reliably, especially for under-represented varieties.

\section{Discussion and Conclusion}
\label{sec:discussion-conclusion}

This study set out to answer two primary research questions:

\noindent \emph{\textbf{(1) Can \textit{ICL} unlock human-like adaptation benefits in a state-of-the-art SLM?}}\\ \textbf{Yes.} Across four corpora that span native, heritage, and second-language English, supplying even a few transcribed examples produces the steep, rapidly-saturating learning curve that psycholinguistic work reports for humans exposed to unfamiliar talkers. Performance gains arise fastest in the first few trials ($\sim$25–30~s of speech), taper off thereafter, and follow two recognizable phases: (i) a \emph{speaker-specific} phase in which 4–6 same-speaker examples yield a $\sim$20\% relative WER reduction, and (ii) a \emph{variety-general} phase in which larger shot counts confer similar benefits even when examples come from different speakers of the same group.

\noindent \emph{\textbf{(2) If so, does this human-like adaptation translate into state-of-the-art recognition across diverse speakers and language varieties?}}\\ \textbf{Largely, but not uniformly.} After 9–12 shots, Phi-4-MM achieves or exceeds state-of-the-art WERs on high-resource American English (<2\%), while delivering sizeable absolute drops (1–3.5 pp.; 10–45\% relative) for low-resource varieties such as Spanish-heritage, L1 Hindi, and L1 Korean.

The asymptotic nature of performance gains, with the largest improvements occurring in the first few examples before plateauing around 6–10 shots, closely parallels psychometric curves observed in human speech perception studies \citep{bradlow_perceptual_2008}. This convergence suggests that ICL accesses fundamental mechanisms of acoustic-phonetic recalibration, potentially involving rapid re-weighting of phonetic features based on observed speaker-specific patterns. The fact that approximately 25–30 seconds of transcribed audio captures most available adaptation benefits mirrors the “fast” adaptation documented in human perceptual learning paradigms.

A notable effect in the results (Table~\ref{tab:main_results}) is the transient increase in WERs for several varieties when transitioning from zero-shot to a single in-context example. We hypothesize that this reflects an initial phase of task adaptation: While Phi-4-MM is instruction-tuned, it was not explicitly trained on ICL for the ASR task, let alone our novel protocol. The first audio-transcript pair may therefore present a dual challenge: the model must first recognize and assimilate the novel task structure itself (essentially learning the "rules" of this ICL interaction) before it can effectively leverage the example's content for acoustic-lexical recalibration towards the target speaker or variety. Once this foundational task understanding is established, subsequent exemplars can more directly contribute to the targeted adaptation, leading to the WER reductions observed with additional examples. Given the strength of this effect in HEC and L2-ARCTIC, there may be interaction effects between task-learning and the out-of-distribution nature of some English varieties, however, analyses across additional corpora would be required to fully disentangle these effects.

The observation that ASR performance is notably sensitive to the prompt offers intriguing parallels to human speech perception. Research by \citet{d2015persona} demonstrates that human listeners' categorization of ambiguous speech sounds is shaped by explicit social information provided about the speaker. The "non-native" prompt appears to prime the ASR system, potentially by activating or re-weighting internal models suited for greater acoustic-phonetic variability or by adjusting decision thresholds, even when such characteristics are not objectively present in the target audio. 

The magnitude of improvements for several low-resource varieties is particularly striking. For example, in CMU-Arctic, the 'Other' subgroup (speakers with Scottish, Canadian, and Israeli backgrounds, as detailed in Section~\ref{subsec:cmuarctic}) achieved a 44.5\% relative WER reduction (0 to 12 shots), and the 'Indian' subgroup saw a 39.2\% reduction. Spanish heritage speakers (HEC) experienced an average relative reduction of 27.2\%, though this figure is significantly influenced by an outlier speaker improving from a very high baseline (see Section~\ref{sec:results}). Other L2-Arctic varieties such as Hindi English (-13.7\%) and Korean English (-7.9\%) also benefited, albeit with more modest relative reductions (Table~\ref{tab:main_results}). These gains help narrow the performance gap compared to high-resource US English. This finding connects to evidence that frequency alone provides limited benefits—our results show that meaningful adaptation requires quality variation, not mere repetition \citep{demarneffe2013interaction}. The disproportionate gains for underrepresented speakers suggest that ICL may help mitigate biases inherent in training distributions dominated by high-resource varieties.

We found that same-speaker examples provide a significant advantage at 4–6 shots (about 1.1 percentage points), but this benefit disappears entirely with longer contexts. This pattern aligns with multi-level adaptation processes in human perception \citep{sidaras_perceptual_2009, moon2013learning}. Adaptation appears to begin with speaker-specific cues and then shift toward variety-level generalization, as the model learns to extract features that transcend individual speaker characteristics.

More importantly, these disproportionate gains for low-resource varieties offer a scalable pathway toward more equitable speech technology, requiring no additional training data or computational resources beyond slightly longer inference contexts. Just as humans cope effortlessly with variation at every level, frontier SLMs show emergent robustness that can be unlocked purely through prompt engineering—offering a practical tool to improve ASR equity for speakers and varieties historically underserved by speech technology. Future work should extend this framework to spontaneous speech, cross-lingual settings, and streaming applications while probing the precise mechanisms of variety-specific adaptation.

\section*{Limitations}
\label{sec:limitations}
While this study demonstrates the significant potential of ICL for ASR adaptation within a frontier model, its limitations define key avenues for future research. 
First, generalizability is constrained: our experiments used a single model (Phi-4-MM) and focused on read English speech. Second, the ICL methodology has scope for expansion. Our reliance on accurately transcribed context, while establishing the potential of ICL with quality signals, may not always be practical. Future efforts should explore unsupervised or self-transcription for context generation and active context selection. Additionally, we explored adaptation mainly within the same speaker or variety and tested limited prompt variations, which inhibited the engineering goal to make speech recognition more robust and restricted the scope of investigations into human-model perceptual convergence.

\section*{Acknowledgments}
We would like to thank Myra Cheng, Aryaman Arora, Martijn Bartelds, Chen Shani, Kaitlyn Zhou, Katia Shutova, Julie Kallini, Ada Tur, and members of the Stanford NLP group for their feedback at various stages of this project.

\bibliography{custom.bib}
\section*{Appendix: Supplementary Experimental Details}

\subsection*{A.1 Model and Generation Parameters}
\label{app:model_params}

\textbf{Model Configuration:}
\begin{itemize}
    \item \textbf{Model:} \texttt{microsoft/Phi-4-multimodal-instruct}
    \item \textbf{Processor:} \texttt{AutoProcessor.from\textunderscore pretrained('microsoft/Phi-4-multimodal-instruct', trust\textunderscore remote\textunderscore code=True)}
    \item \textbf{Loading:} \texttt{AutoModelForCausalLM.from\textunderscore pretrained('microsoft/Phi-4-multimodal-instruct', trust\textunderscore remote\textunderscore code=True, torch\textunderscore dtype='auto', attn\textunderscore implementation='flash\textunderscore attention\textunderscore 2')}
\end{itemize}

\textbf{Generation Configuration:}
\begin{itemize}
    \item \texttt{max\textunderscore new\textunderscore tokens}: 1200
    \item \texttt{do\textunderscore sample}: False (greedy decoding)
    \item \texttt{num\textunderscore beams}: 1
    \item \texttt{num\textunderscore logits\textunderscore to\textunderscore keep}: 1 (explicitly set at multiple levels)
\end{itemize}

\subsection*{A.2 Datasets and Comprehensive Preprocessing}
\label{app:datasets_preprocessing}

\subsubsection*{A.2.1 Dataset Sources}
\begin{itemize}
    \item \textbf{L2-Arctic:} \texttt{NathanRoll/l2-arctic-dataset-250}
    \item \textbf{HEC (HISP-ENG):} \texttt{NathanRoll/hisp-eng}
    \item \textbf{CMU-Arctic:} \texttt{NathanRoll/cmu-arctic}
\end{itemize}

\subsubsection*{A.2.2 Audio Preprocessing Pipeline}

\textbf{Target Specifications:}
\begin{itemize}
    \item Sample Rate: 16,000 Hz (resampled using \texttt{librosa.resample})
    \item Format: 32-bit float (\texttt{np.float32})
    \item Duration: Minimum 2.5 seconds (shorter clips filtered out)
\end{itemize}

\textbf{Normalization Algorithm Steps:}
\begin{itemize}
    \item \textbf{Integer handling:} Convert integer types by dividing by max value for that dtype
    \item \textbf{Float handling:} Convert directly to \texttt{np.float32}
    \item \textbf{FLAC bug detection:} Check for max > 0.99 and min > -0.5, indicating missing negative values
    \item \textbf{Bug correction:} Flip values above 0.9 threshold if bug detected
    \item \textbf{Range clipping:} Clip extreme values exceeding $\pm$1.1 to [-1.0, 1.0]
\end{itemize}

\textbf{Resampling Configuration:}
\begin{itemize}
    \item Primary method: \texttt{librosa.resample} with default parameters
    \item Fallback method: \texttt{librosa.resample} with \texttt{res\_type='kaiser\textunderscore fast'} if primary fails
    \item All resampling errors are logged and re-raised for debugging
\end{itemize}

\subsubsection*{A.2.3 Text Normalization}

\textbf{Normalization Steps:}
\begin{enumerate}
    \item Convert to lowercase
    \item Remove punctuation: \texttt{. , ? ! ; : " ' ( ) [ ]} (each replaced with space)
    \item Normalize whitespace: Multiple spaces collapsed to single spaces, leading/trailing whitespace removed
\end{enumerate}

\textbf{Implementation Logic:}
Convert text to lowercase, iterate through punctuation list replacing each with space, then split and rejoin to normalize whitespace.

\subsubsection*{A.2.4 Dataset Filtering Criteria}

\textbf{Variety-Level Filtering:}
\begin{itemize}
    \item Varieties must have $\geq$2 speakers
    \item For HEC dataset: exclude samples with \texttt{variety == 'unknown'}
\end{itemize}

\textbf{Speaker-Level Filtering:}
\begin{itemize}
    \item Speakers must have $\geq$(max\textunderscore shots + 1) valid utterances
    \item Maximum 50 samples per speaker used (selected via shuffling with speaker-specific seed)
\end{itemize}

\textbf{Sample-Level Filtering:}
\begin{itemize}
    \item Duration $\geq$2.5 seconds
    \item Valid audio array present
    \item Valid transcript field present and non-empty
\end{itemize}

\textbf{Variety Mapping Details:}
\begin{itemize}
    \item \textbf{CMU-Arctic:} Based on speaker ID mapping to variety (see \texttt{CMU\textunderscore ARCTIC\textunderscore VARIETIES})
    \item \textbf{HEC:} Based on speaker origin mapping (see \texttt{HISP\textunderscore ENG\textunderscore ORIGINS})
    \item \textbf{L2-Arctic:} Uses \texttt{l1} field directly
\end{itemize}

\subsection*{A.3 Experimental Design and Context Selection}
\label{app:experimental_design}

\subsubsection*{A.3.1 Random Seed Management}

\textbf{Global Seed:} 42 (default, configurable via command line)

\textbf{Seed Hierarchies:}
\begin{itemize}
    \item \textbf{Speaker-level shuffling:} \texttt{global\_seed + hash(f"\{variety\}\textunderscore \{speaker\}") \% 10000}
    \item \textbf{Trial-level context selection:} \texttt{global\_seed + hash(f"\{speaker\}\textunderscore \{trial\textunderscore idx\}") \% 10000}
    \item \textbf{Different-speaker selection:} Same as trial-level but includes variety information
\end{itemize}

This hierarchical seeding ensures:
\begin{enumerate}
    \item Reproducible speaker orderings
    \item Consistent context selection across runs
    \item Deterministic different-speaker selection
\end{enumerate}

\subsubsection*{A.3.2 Context Example Selection Algorithm}

\textbf{Same-Speaker Condition Logic:}
\begin{itemize}
    \item Build candidate list excluding current test sample
    \item Filter out samples with identical normalized transcripts 
    \item Use trial-specific random seed for selection
    \item Sample n\textunderscore shots examples without replacement
\end{itemize}

\textbf{Different-Speaker Condition Logic:}
\begin{itemize}
    \item Select random other speaker from same variety
    \item Collect samples from selected speaker
    \item Filter out samples with identical transcripts to test audio
    \item Sample n\textunderscore shots examples using same random seed strategy
\end{itemize}

\subsubsection*{A.3.3 Trial Generation Process}

\textbf{Trial Count Calculation:}
\begin{itemize}
    \item Maximum trials per speaker: min(pool\textunderscore size - n\textunderscore shots, max\textunderscore trials)
    \item Pool size limit: 50 samples per speaker
    \item Test samples drawn sequentially from shuffled pool
\end{itemize}

\textbf{Quality Control:}
\begin{itemize}
    \item Skip trials where insufficient context examples available
    \item Skip samples without valid transcript fields
    \item Handle all exceptions gracefully with detailed logging
\end{itemize}

\subsection*{A.4 In-Context Learning Prompts}
\label{app:icl_prompts}

The following prompt structures are used for the Phi-4 model, where \texttt{<|user|>}, \texttt{<|assistant|>}, \texttt{<|audio\textunderscore N|>}, and \texttt{<|end|>} are special model tokens.

\subsubsection*{A.4.1 Zero-Shot Prompts}

\textbf{Standard Prompt:}
\texttt{<|user|><|audio\textunderscore 1|>Transcribe the audio clip into text.<|end|><|assistant|>}

\textbf{Variation Prompt (Non-Native Focus):}
\texttt{<|user|><|audio\textunderscore 1|>Transcribe the audio clip from a non-native English speaker into text.<|end|><|assistant|>}

\subsubsection*{A.4.2 Few-Shot Prompt Structure}

\textbf{Initial Instruction Block:}
\begin{itemize}
    \item User message: Explains providing N examples from non-native speaker, followed by new audio from same/different speaker
    \item Assistant acknowledgment: Confirms understanding and intent to use examples for transcription
\end{itemize}

\textbf{Dynamic Elements:}
\begin{itemize}
    \item \texttt{\{num\textunderscore shots\textunderscore text\}}: ``an example'' (1-shot) or ``N examples'' (N-shot)
    \item \texttt{\{speaker\textunderscore reference\}}: ``the same speaker'' or ``a different speaker''
    \item \texttt{\{pronoun\textunderscore text\}}: ``it'' (1-shot) or ``them'' (N-shot)
\end{itemize}

\textbf{Example Block Structure:}
\begin{itemize}
    \item \textbf{Standard:} User provides audio, assistant responds with transcript
    \item \textbf{Variation:} Same as standard but assistant response prefixed with ``Transcription: ''
\end{itemize}

\textbf{Final Query Block:}
User provides final audio with speaker reference, assistant begins response (with ``Transcription: '' prefix for variation prompt).

\subsection*{A.5 Computational Requirements and Implementation}
\label{app:computational}

\textbf{Hardware Specifications:}
\begin{itemize}
    \item GPU: NVIDIA A100 (required for flash attention)
    \item Memory: Minimum 40GB GPU memory recommended
    \item Runtime: 8-12 hours per corpus for complete evaluation (0-12 shots)
\end{itemize}

\textbf{Software Dependencies:}
\begin{itemize}
    \item torch, peft, torchvision, backoff, flash-attn
    \item tqdm, jiwer, librosa, transformers, datasets
\end{itemize}

\textbf{Model Memory Management:}
\begin{itemize}
    \item Model loaded with \texttt{torch\textunderscore dtype='auto'} for optimal precision/memory trade-off
    \item Flash attention implementation used to reduce memory footprint
    \item Single inference batch processing (no batching across utterances)
\end{itemize}

\subsection*{A.6 Statistical Analysis and Data Collection}
\label{app:statistical}

\subsubsection*{A.6.1 Trial Collection}

\textbf{Number of Trials:}
\begin{itemize}
    \item Default maximum: 50 trials per speaker per condition
    \item Actual trials: min(available\_samples - n\textunderscore shots, max\_trials)
    \item Zero-shot: All valid samples used (up to 50)
\end{itemize}

\textbf{Data Validation:}
\begin{itemize}
    \item WER calculated using \texttt{jiwer} library with normalized texts
    \item All results stored with full precision (no rounding during intermediate calculations)
    \item Individual trial results preserved in addition to averages
\end{itemize}

\subsubsection*{A.6.2 Result Aggregation}

\textbf{Speaker-Level Results Format:}
\begin{itemize}
    \item Variety identification
    \item Run counts per shot condition
    \item Average WER per shot condition
    \item Complete list of individual WER values
\end{itemize}

\textbf{Corpus-Level Results:}
\begin{itemize}
    \item Weighted averages across all speakers
    \item Total sample counts per condition
    \item Preservation of speaker-level breakdowns
\end{itemize}

\subsection*{A.7 Reproducibility Checklist}
\label{app:reproducibility}

\textbf{1. Environment Setup:}
\begin{itemize}
    \item Use identical package versions (see dependencies list)
    \item Set all random seeds (Python, NumPy, PyTorch)
\end{itemize}

\textbf{2. Data Processing:}
\begin{itemize}
    \item Apply exact audio normalization pipeline (including FLAC bug correction)
    \item Use identical text normalization (case, punctuation, whitespace)
    \item Apply same filtering criteria (duration, variety, speaker counts)
\end{itemize}

\textbf{3. Experimental Configuration:}
\begin{itemize}
    \item Use hierarchical random seeding as specified
    \item Maintain exact prompt structure (including special tokens)
    \item Follow context selection algorithm precisely
\end{itemize}

\textbf{4. Model Configuration:}
\begin{itemize}
    \item Use greedy decoding (do\_sample=False)
    \item Set num\textunderscore logits\textunderscore to\textunderscore keep=1 at all levels
    \item Use flash attention implementation
\end{itemize}

\textbf{5. Evaluation:}
\begin{itemize}
    \item Calculate WER using jiwer with normalized texts
    \item Aggregate results maintaining full precision
    \item Store individual trial values, not just averages
\end{itemize}

\subsection*{A.8 Extended Speaker-Level Results}
\label{app:speaker_results}

The following table provides complete speaker-level results for 0-shot and 12-shot conditions across all corpora, enabling verification of reported aggregate statistics.

\begin{table*}[htbp]
\centering
\caption{Complete speaker-level Word Error Rates (WER) for 0-shot and 12-shot conditions. All WERs are percentages.}
\label{tab:complete_speaker_results}
\begin{tabular}{llcc}
\toprule
Dataset (Corpus) & Speaker & 0-shot WER (\%) & 12-shot WER (\%) \\
\midrule
CMU-Arctic & aew & 1.5 & 0.7 \\
CMU-Arctic & ahw & 2.8 & 2.0 \\
CMU-Arctic & aup & 3.1 & 2.1 \\
CMU-Arctic & axb & 3.8 & 2.4 \\
CMU-Arctic & bdl & 1.4 & 0.9 \\
CMU-Arctic & clb & 1.9 & 0.7 \\
CMU-Arctic & eey & 1.9 & 2.6 \\
CMU-Arctic & fem & 4.1 & 2.5 \\
CMU-Arctic & gka & 1.9 & 1.3 \\
CMU-Arctic & ksp & 3.4 & 2.4 \\
CMU-Arctic & ljm & 2.2 & 1.8 \\
CMU-Arctic & lnh & 2.2 & 1.0 \\
CMU-Arctic & rms & 2.0 & 0.7 \\
CMU-Arctic & slp & 3.9 & 1.8 \\
CMU-Arctic & slt & 1.7 & 1.1 \\
HEC & 0 & 4.6 & 7.6 \\
HEC & 1 & 3.9 & 3.3 \\
HEC & 18 & 6.7 & 4.9 \\
HEC & 3 & 2.3 & 1.9 \\
HEC & 4 & 4.7 & 3.8 \\
HEC & 6 & 3.0 & 2.3 \\
HEC & 7 & 63.9 & 41.0 \\
L2-Arctic & ASI & 3.5 & 3.3 \\
L2-Arctic & BWC & 9.4 & 10.3 \\
L2-Arctic & EBVS & 8.8 & 7.5 \\
L2-Arctic & ERMS & 6.5 & 6.1 \\
L2-Arctic & HJK & 4.1 & 4.0 \\
L2-Arctic & HQTV & 17.9 & 15.8 \\
L2-Arctic & LXC & 6.4 & 6.1 \\
L2-Arctic & MBMPS & 5.5 & 5.1 \\
L2-Arctic & NCC & 6.4 & 6.7 \\
L2-Arctic & NJS & 5.1 & 4.9 \\
L2-Arctic & PNV & 4.6 & 5.6 \\
L2-Arctic & RRBI & 4.9 & 3.5 \\
L2-Arctic & TNI & 3.7 & 3.5 \\
L2-Arctic & YKWK & 4.2 & 3.7 \\
\midrule
\multicolumn{2}{l}{\textbf{Grand Average (across speakers)}} & \textbf{6.1} & \textbf{4.9} \\
\bottomrule
\end{tabular}
\end{table*}

\end{document}